\documentclass{article}
\usepackage{iclr2027_conference,times}

\usepackage{amsmath,amsfonts,bm}

\def\Figref#1{Figure~\ref{#1}}

\def\eqref#1{equation~\ref{#1}}
\def\Eqref#1{Equation~\ref{#1}}
\def\1{\bm{1}}

\DeclareMathAlphabet{\mathsfit}{\encodingdefault}{\sfdefault}{m}{sl}
\SetMathAlphabet{\mathsfit}{bold}{\encodingdefault}{\sfdefault}{bx}{n}

\usepackage[hidelinks]{hyperref}
\usepackage{url}

\usepackage{amsmath}
\usepackage{amssymb}
\usepackage{booktabs}
\usepackage{graphicx}
\usepackage{wrapfig}
\usepackage{kotex}

\definecolor{Rhodamine}{RGB}{227, 0, 122}

\title{Policy-Aligned Predictive Verification for Adaptive VLA Action-Chunk Execution}

\author{%
\textbf{Minkyu Kim}$^{1}$ \quad \textbf{Junhoo Lee}$^{2}$ \quad \textbf{Seungyeon Kim}$^{1}$ \quad \textbf{Suhyun Jeon}$^{1}$ \\
\textbf{Baekseung Kim}$^{1}$ \quad \textbf{Donghoon Han}$^{3}$ \quad \textbf{Nojun Kwak}$^{1}$ \\
{\normalfont $^{1}$Seoul National University \quad $^{2}$KAIST \quad $^{3}$Dnotitia}
}

\iclrfinalcopy
\begin{document}

\maketitle
\lhead{Preprint}

\begin{abstract}
Vision-language-action (VLA) policies generate action sequences that reduce repeated inference, but deciding how much of each sequence to execute is critical to task success. Existing adaptive methods make this decision using generation stability or prediction agreement, yet these signals do not establish whether executing a candidate prefix preserves the ability to complete the task. We introduce a \emph{policy-aligned predictive verifier} (PAV) that anticipates the consequences of candidate action prefixes and evaluates their compatibility with successful continuation before execution. Rather than using future prediction to generate actions, our method evaluates existing proposals in the VLA's internal task-conditioned representation space. To learn local judgments from terminal task outcomes, we combine trajectory-level survival supervision with a localization prior that preserves useful behavior within failed rollouts. The resulting validity scores guide execution-horizon selection. Evaluations with $\pi_0$ and $\pi_{0.5}$ across LIBERO, RoboTwin 2.0, and real-world manipulation show improved success--query trade-offs. These findings point toward a different role for future prediction in robotics, using it to assess how far to trust a policy's proposed actions rather than to generate new ones.
\end{abstract}

\section{Introduction}

\begin{wrapfigure}{r}{0.56\textwidth}
  \vspace{-\baselineskip}
  \centering
  \setlength{\abovecaptionskip}{4pt}
  \includegraphics[width=\linewidth]{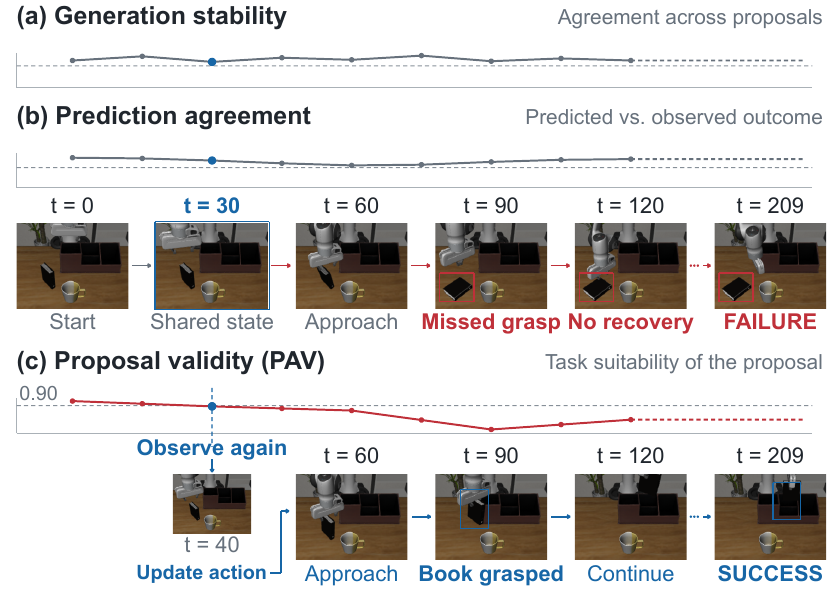}
  \caption{Generation stability (a) and prediction agreement (b) can remain high despite a missed grasp and task failure. Proposal validity (c) guides earlier replanning, leading to success in this example.}
  \label{fig:motivation}
  \vspace{-1pt}
\end{wrapfigure}

Vision-Language-Action (VLA) models have shown strong performance in generating robot actions from visual observations and language instructions \citep{brohan2023rt,kim2024openvla,black2024pi_0,miao2026jepa}. Yet querying a large VLA policy at every control step is costly, while robot behavior requires coordinated motion over multiple steps. Action chunking addresses both demands by generating a sequence of future actions in one query, supporting temporal coherence with fewer policy calls. However, executing an entire chunk without new observations limits responsiveness to environmental changes and allows execution errors to accumulate. Many systems therefore use a shorter \emph{execution horizon} to execute only an initial prefix of each chunk before replanning from new observations. Adapting this horizon is thus central to retaining the efficiency of action chunking while remaining responsive to feedback.

Adaptive execution seeks the longest prefix of each action chunk that still allows the policy to complete the task successfully. Identifying this prefix requires anticipating how the proposed actions will affect the policy's ability to complete the remaining task. Because the effects of a candidate prefix on task success are not directly observable before execution, existing approaches rely on indirect signals to guide execution. Some methods use generation stability as this indirect signal, drawing on internal policy information to select an execution horizon \citep{wang2026vla,feng2026denoising,chen2026dynamic}. Others use prediction agreement, comparing action-conditioned predictions with incoming observations to guide correction or replanning during execution \citep{pan2026vla,liu2026checkvla}. Generation stability and prediction agreement alone, however, do not establish whether a candidate prefix preserves compatibility with successful task completion. As illustrated in \Figref{fig:motivation}, an action proposal can be consistently generated and its consequences accurately predicted, yet executing it can still lead to a missed grasp and task failure.

To assess this compatibility directly, we introduce \emph{proposal validity}, which estimates the probability that the policy can successfully complete the task after executing a candidate prefix. This probability depends on the state reached after executing the prefix, because the policy must complete the remaining task from that state. We therefore predict this future state to evaluate whether it supports successful continuation under the policy. World models and World Action Models have demonstrated this capability, building on pretrained visual representations or video models to predict future observations or latent states for planning and action generation \citep{seo2022reinforcement,du2023learning,zhou2024dino,assran2025v,ye2026world,sun2026vla}. However, because VLA policies already generate effective action proposals, we use this predictive capability to evaluate those proposals rather than to generate new actions. To ground this evaluation in the same information used for action generation, we predict the effects of each prefix in the VLA's task-conditioned representation space, which integrates visual observations and language instructions.

We therefore propose a \emph{policy-aligned predictive verifier} (PAV) that assesses each action extension's compatibility with successful continuation from predicted VLA representations. These extension-level assessments provide local validity estimates used to select execution horizons. To learn these local judgments, however, we must rely on rollouts that provide only terminal task outcomes. These outcomes do not reveal the validity of individual extensions, since even a failed rollout may contain valid action prefixes. We therefore treat extension-level validity as latent and supervise its accumulated product using the terminal outcome. However, the model can fit a failed rollout by slightly lowering validity scores across the rollout, including those of valid extensions. We therefore introduce a localization prior that encourages the model to concentrate these reductions in fewer extensions. At inference time, we combine the local validity scores to evaluate each candidate prefix, accounting for its length. We use these prefix-level scores to decide how many actions to execute before obtaining a new observation and querying the policy again.

Our experiments with $\pi_0$ and $\pi_{0.5}$ on LIBERO, RoboTwin 2.0, and real-world tasks assess whether proposal validity selects execution horizons that improve success with fewer policy queries. To examine the basis for these decisions, prediction tests assess action sensitivity, while ablations examine future prediction, policy alignment, and the contributions of individual loss terms. Predicting future VLA representations achieves higher success with fewer queries than direct verification or prediction in generic visual features.

\section{Related Work}
\label{sec:related-work}

\subsection{Adaptive Execution of Action Chunks}

Action-sequence policies such as ACT and $\pi_0$ generate future action chunks \citep{zhao2023learning,black2024pi_0}. A fixed execution horizon uses the same prefix length across proposals and states, trading fewer queries for later feedback. Adaptive execution instead adjusts when to request a new proposal. One line of work uses internal policy signals to assess generation stability and guide horizon selection. AutoHorizon estimates an execution limit from action self-attention \citep{wang2026vla}, while $A^3$ selects an execution horizon using action consensus and re-decoding stability \citep{chen2026dynamic}. Another line of work assesses prediction agreement from observations during execution. VLA-Corrector truncates and replans when predicted and observed visual dynamics persistently diverge, whereas CheckVLA relates prediction-observation discrepancies to failure risk and latency-aware suffix repair \citep{pan2026vla,liu2026checkvla}. These approaches use generation stability or prediction agreement to guide execution, whereas our verifier predicts the consequences of candidate prefixes and evaluates their compatibility with successful continuation before execution. To our knowledge, we are the first to combine prefix-conditioned future prediction with proposal-validity estimation for adaptive action-chunk execution.

\subsection{World Models and Predictive Representations}

World models capture action-conditioned dynamics over future observations or latent states \citep{chua2018deep,hafner2019learning,hansen2022temporal,hansen2024td}. Dreamer learns behavior from imagined latent trajectories \citep{hafner2019dream}, DINO-WM predicts pretrained visual features for goal-directed planning \citep{zhou2024dino}, and World Action Models such as DreamZero jointly predict future video and actions \citep{ye2026world}. Value Prediction Networks evaluate imagined futures through rewards and values \citep{oh2017value}, while Diffuser generates trajectories for planning \citep{janner2022planning}. Beyond these uses, prediction can support proposal verification. Given a VLA proposal, future prediction anticipates the consequences of executing candidate prefixes. Assessing these consequences requires representations that retain information relevant to control. DeepMDP and bisimulation-based learning connect representations to rewards and transitions \citep{gelada2019deepmdp,zhang2020learning}, while VIP learns visual representations through goal-conditioned value learning \citep{ma2022vip}. We therefore predict prefix-induced futures in the action-generating policy's internal representation space, retaining the joint observation and instruction context for outcome-based proposal verification.

\subsection{Learning from Trajectory-Level Outcomes}
\label{sec:credit-assignment}

Learning prefix-level validity from terminal outcomes is related to multiple-instance learning, where collection-level labels leave individual instance labels unobserved \citep{dietterich1997solving,ilse2018attention}. Here, rollout outcomes supervise temporally ordered prefix extensions without identifying which extension caused a failure. Reward redistribution similarly derives local learning signals from delayed returns or demonstrations \citep{arjona2019rudder,ren2021learning,patil2020align}. A related credit-assignment problem arises in language-model reinforcement learning \citep{chan2024dense}, where RED derives token-level rewards from holistic response feedback \citep{li2025red}. These approaches recover local information from global outcomes \citep{yuan2024free}, but we use this supervision to verify proposed execution and select horizons rather than optimize the action-generating policy.

\section{Method}
\label{sec:method}

\subsection{Preliminaries}
\label{sec:preliminaries}

\paragraph{Action-chunk execution.}
Adaptive action-chunk execution aims to select the longest prefix likely to preserve compatibility with successful continuation. Given observation $o_t$ and instruction $\ell$ at decision time $t$, policy $\pi$ generates a length-$H$ chunk \citep{zhao2023learning,black2024pi_0}. With $\mathbf a_{t,j}$ its $j$-th action, choosing an execution horizon $h$ from $[H]\triangleq\{1,\ldots,H\}$ commits to the prefix
\begin{equation}
  \mathbf A_t\sim\pi(\,\cdot\mid o_t,\ell),
  \qquad
  \mathbf A_t^{1:h}=(\mathbf a_{t,1},\ldots,\mathbf a_{t,h}),
  \qquad h\in[H].
  \label{eq:action-chunk}
\end{equation}
After $h$ control steps, the policy receives $o_{t+h}$ and replans. Choosing $h$ therefore balances inference cost and temporal consistency against responsiveness \citep{wang2026vla}.

\paragraph{Generation stability as an execution criterion.}
Because future execution outcomes are unavailable when choosing a horizon, $A^3$ \citep{chen2026dynamic} tests whether proposed actions remain stable when other actions in the chunk are fixed as context. Let $\widetilde{\mathbf a}_{t,j}$ denote the action regenerated under such a context. This generation-stability signal can be expressed conceptually through the discrepancy
\begin{equation}
  e_t(j)=\left\|\widetilde{\mathbf a}_{t,j}-\mathbf a_{t,j}\right\|_2^2.
  \label{eq:generation-stability}
\end{equation}
A small discrepancy indicates that the policy reproduces the proposed action under the imposed context. $A^3$ therefore accepts the longest prefix that passes its conditional consistency checks. However, consistently generated actions can still prevent later task completion, so generation stability does not establish compatibility with successful continuation.

\paragraph{Prediction agreement as an execution criterion.}
Prediction agreement provides evidence for intervention by comparing action-conditioned predictions with observations collected during execution \citep{pan2026vla,liu2026checkvla}. To express this signal generically, let $x_{t'}$ denote an observed feature statistic at control step $t'>t$, $\hat x_{t'}$ its earlier prediction, and $d$ a discrepancy measure, giving
\begin{equation}
  d_{t'}=d\!\left(\hat x_{t'},x_{t'}\right).
  \label{eq:prediction-discrepancy}
\end{equation}
A larger discrepancy indicates weaker prediction agreement and provides evidence for correction or replanning during execution. Our decision instead requires evaluating the compatibility of a candidate prefix before the corresponding observations become available.

\subsection{Formulating Proposal Validity}
\label{sec:predictive-verification}

\paragraph{Defining the decision target.}
We therefore evaluate candidate prefixes by their compatibility with successful continuation rather than by generation stability or prediction agreement alone. \emph{Proposal validity} quantifies this compatibility as the probability that successful continuation remains possible throughout the proposed execution. To assess the whole prefix, let $\tau_t^{1:h}$ denote the trajectory induced by the first $h$ actions of a fixed proposal $\mathbf A_t$, and let $\mathcal T_\ell^{\pi,+}$ contain trajectories from which successful continuation under $\pi$ remains possible after every extension. We include task progress and remaining time in the state to assess that possibility, and hold the continuation protocol constant to isolate the consequences of the candidate prefixes. Formally,
\begin{equation}
  V_t(h)
  :=\Pr\!\left(
    \tau_t^{1:h}\in\mathcal T_\ell^{\pi,+}
    \,\middle|\,o_t,\ell,\mathbf A_t
  \right).
  \label{eq:proposal-validity}
\end{equation}
Requiring compatibility throughout execution makes $\mathcal T_\ell^{\pi,+}$ prefix-closed, so a compatible trajectory has compatible shorter prefixes. This gives $V_t(h)$ the cumulative structure needed to judge how far execution can proceed while preserving compatibility.

To reduce queries while preserving compatibility, an oracle would select the longest prefix with sufficiently high validity. Let $\mathcal H\subseteq[H]$ denote the candidate execution horizons and $\epsilon\in(0,1)$ a small tolerance. The target horizon is
\begin{equation}
  h_t^{\mathrm{target}}
  =\max\!\left(\{\min\mathcal H\}\cup
    \{h\in\mathcal H\mid V_t(h)\ge 1-\epsilon\}\right).
  \label{eq:target-horizon}
\end{equation}
\paragraph{Decomposing proposal validity.}
To estimate validity before executing progressively longer prefixes, we separate the conditional compatibility of each additional step from the validity accumulated over preceding steps. Since a compatible prefix has compatible predecessors,
\begin{equation}
  V_t(h)
  =V_t(h-1)\,
    \Pr\!\left(
      \tau_t^{1:h}\in\mathcal T_\ell^{\pi,+}
      \,\middle|\,
      \tau_t^{1:h-1}\in\mathcal T_\ell^{\pi,+},
      o_t,\ell,\mathbf A_t
    \right).
  \label{eq:validity-recursion}
\end{equation}
We set $V_t(0)=1$ because no proposed action has yet been executed, and consider the conditional factor where $V_t(h-1)>0$. Assessing this factor before execution requires anticipating the extension's consequence. We therefore marginalize over its future policy representation $z_{t+h}$ to separate prediction from verification,
\begin{equation}
  \begin{aligned}
  V_t(h)={}&V_t(h-1)\int
    \underbrace{p\!\left(z\mid
      \tau_t^{1:h-1}\in\mathcal T_\ell^{\pi,+},o_t,\ell,\mathbf A_t
    \right)}_{\text{prediction}}\\
    &\quad\times\underbrace{\Pr\!\left(
      \tau_t^{1:h}\in\mathcal T_\ell^{\pi,+}
      \mid z_{t+h}=z,
      \tau_t^{1:h-1}\in\mathcal T_\ell^{\pi,+},o_t,\ell,\mathbf A_t
    \right)}_{\text{verification}}
    \,\mathrm dz.
  \end{aligned}
  \label{eq:value-decomposition}
\end{equation}
Here $p$ denotes the conditional density of $z_{t+h}$. This decomposition motivates a predictor trained on observed futures and a verifier trained on task outcomes.

\begin{figure}[t]
  \centering
  \includegraphics[width=\linewidth]{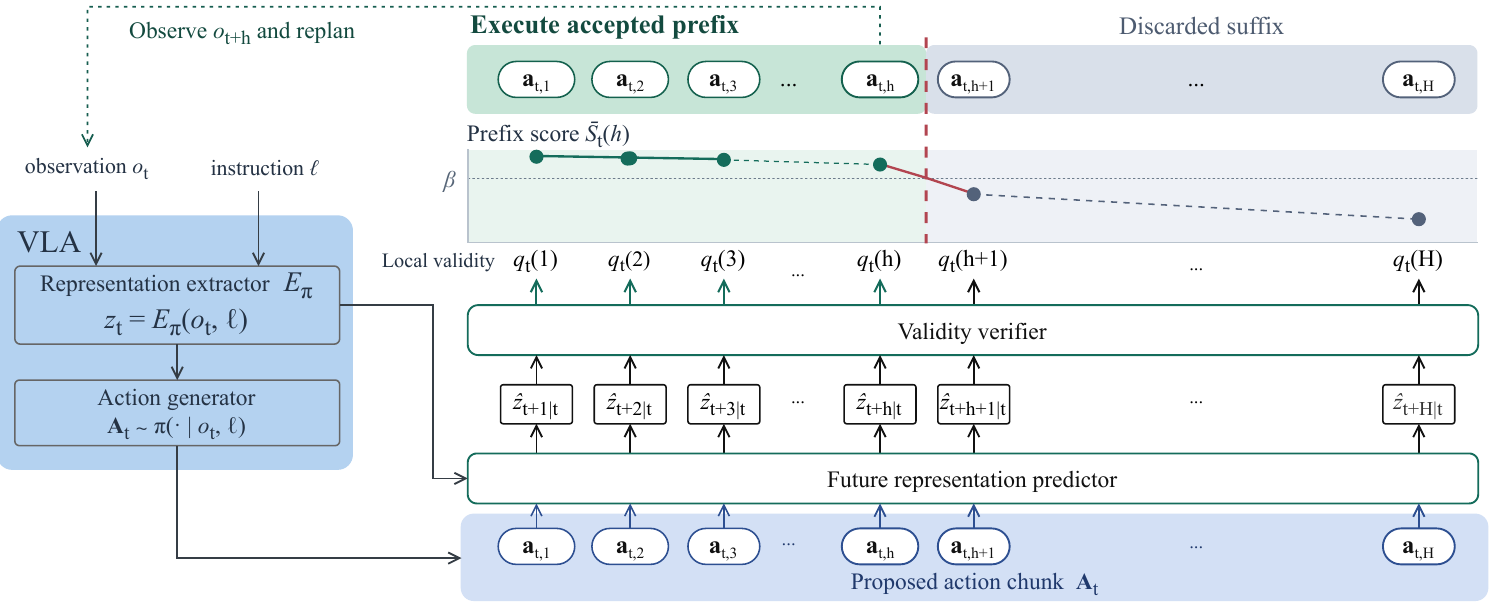}
  \caption{Policy-aligned predictive verification. The predictor uses the VLA representation and proposed actions to predict future policy representations. The verifier estimates local validity, which is accumulated and normalized to select an execution horizon. The robot executes the selected prefix and replans.}
  \label{fig:overview}
\end{figure}

\subsection{Learning Prediction and Verification}
\label{sec:trajectory-supervision}

\paragraph{Predicting action consequences in policy space.}
To approximate the prediction factor in \Eqref{eq:value-decomposition}, we use a learned point prediction $\hat z_{t+h\mid t}$ for verification, as illustrated in \Figref{fig:overview}. To retain the task context used for action selection, we predict in the policy's representation space, with $z_t=E_\pi(o_t,\ell)$ denoting the features extracted by $E_\pi$ from the observation and instruction. Encoding observations after executed prefixes supplies targets for the predictor $f_\theta$,
\begin{equation}
  z_{t+h}=E_\pi(o_{t+h},\ell),
  \qquad
  \hat z_{t+h\mid t}=f_\theta(z_t,\mathbf A_t^{1:h}).
  \label{eq:latent-prediction}
\end{equation}

Following the use of observed future representations as prediction targets \citep{schwarzer2020data}, we align our predictions with targets from executed transitions $(o_t,\ell,\mathbf A_t^{1:h},o_{t+h})$ collected in $\mathcal D_{\mathrm{trans}}$. We use cosine distance $d_{\cos}(u,v)=1-\operatorname{sim}(u,v)$, where $\operatorname{sim}$ is cosine similarity. Alignment alone may allow the predictor to ignore which actions produced the future. We therefore pair transitions with alternative prefixes $\tilde{\mathbf A}_t^{1:h}$ in $\mathcal D_{\mathrm{swap}}$, giving $\tilde z_{t+h\mid t}=f_\theta(z_t,\tilde{\mathbf A}_t^{1:h})$. Requiring the executed prefix's predicted representation to be closer to the target by margin $m>0$ encourages action sensitivity without assigning the alternative a failure label. With nonnegative weights $\lambda_{\mathrm{dyn}},\lambda_{\mathrm{swap}}$ and $[x]_+=\max(x,0)$, these requirements give
\begin{equation}
  \begin{aligned}
    \mathcal L_{\mathrm{pred}}(\theta)
      ={}&\lambda_{\mathrm{dyn}}\,
        \mathbb E_{\mathcal D_{\mathrm{trans}}}
        \bigl[d_{\cos}(\hat z_{t+h\mid t},z_{t+h})\bigr]\\
      &+\lambda_{\mathrm{swap}}\,
        \mathbb E_{\mathcal D_{\mathrm{swap}}}
        \Bigl[\bigl[m+d_{\cos}(\hat z_{t+h\mid t},z_{t+h})
          -d_{\cos}(\tilde z_{t+h\mid t},z_{t+h})\bigr]_+\Bigr].
  \end{aligned}
  \label{eq:prediction-objective}
\end{equation}
\providecommand{\notreported}{\textemdash}
\begin{table*}[t]
\centering
\setlength{\belowcaptionskip}{5pt}
\caption{LIBERO success rates (\%) and mean VLA queries per episode. Our method and fixed-horizon baselines use 50 trials per task. $^\dagger$ denotes published reference results with potentially different protocols. Bold and underlining indicate the highest and second-highest success rates within each policy setting, including ties. Dashes indicate unreported results.}
\label{tab:libero_results}
\small
\renewcommand{\arraystretch}{1.08}
\setlength{\arrayrulewidth}{0.3pt}
\setlength{\aboverulesep}{0pt}
\setlength{\belowrulesep}{0pt}
\setlength{\tabcolsep}{6pt}
\newcommand{\liberoheader}{%
  Method & Spatial $\uparrow$ & Object $\uparrow$ & Goal $\uparrow$ &
  Long $\uparrow$ & Avg. $\uparrow$ & Queries $\downarrow$}
\begin{tabular}{@{}l|ccccc|c@{}}
\midrule[0.9pt]
\multicolumn{7}{c}{\rule[-4pt]{0pt}{14pt}\normalsize $\pi_0$\enspace($H=50$)} \\
\midrule[0.9pt]
\liberoheader \\
\midrule
$h=5$ (Baseline) & \textbf{97.4} & 97.2 & 91.0 & 82.8 & 92.10 & 34.6 \\
$h=10$ & 96.0 & \underline{98.4} & \underline{93.6} & 84.2 & 93.05 & 17.2 \\
$h=30$ & 89.4 & 94.4 & 90.2 & 86.6 & 90.15 & 6.3 \\
$h=50$ & 76.6 & 75.4 & 83.4 & 68.8 & 76.05 & 4.4 \\
\midrule
AutoHorizon & 95.4 & \underline{98.4} & \textbf{94.8} & \textbf{87.6} & \underline{94.05} & 10.6 \\
% DVAC, Appendix Tables 7 and 8, pi0 DVAC alpha=2.
DVAC$^\dagger$ & 65.0 & 76.0 & 76.0 & 38.0 & 63.75 & 6.7 \\
\textbf{PAV (Ours)} & \underline{96.4} & \textbf{98.8} & \textbf{94.8} & \underline{87.2} & \textbf{94.30} & 12.3 \\
\midrule[0.9pt]
\multicolumn{7}{c}{\rule[-4pt]{0pt}{14pt}\normalsize $\pi_{0.5}$\enspace($H=10$)} \\
\midrule[0.9pt]
\liberoheader \\
\midrule
$h=5$ (Baseline) & 98.6 & 99.0 & \underline{98.2} & 91.4 & 96.80 & 32.3 \\
$h=10$ & 98.0 & \underline{99.6} & \underline{98.2} & 94.2 & 97.50 & 16.1 \\
\midrule
AutoHorizon & \underline{99.8} & \underline{99.6} & 97.2 & 94.6 & 97.8 & 27.8 \\
% DVAC, Tables 4 and 8, pi0.5 DVAC alpha=2.
DVAC$^\dagger$ & 99.0 & 98.0 & 98.0 & \textbf{97.0} & \underline{98.00} & 18.6 \\
$A^3$ & 99.2 & \underline{99.6} & 97.4 & 94.0 & 97.55 & 16.4 \\
% VLA-Corrector, Table 2, few-shot pi0.5. The source orders Object before Spatial.
VLA-Corrector$^\dagger$ & \textbf{100.0} & \textbf{99.8} & 98.0 & 93.4 & 97.80 & \notreported \\
\textbf{PAV (Ours)} & 99.0 & 99.4 & \textbf{98.4} & \underline{95.6} & \textbf{98.10} & 16.8 \\
\midrule[0.9pt]
\multicolumn{7}{c}{\rule[-4pt]{0pt}{14pt}\normalsize $\pi_{0.5}$\enspace($H=50$)} \\
\midrule[0.9pt]
\liberoheader \\
\midrule
$h=5$ & 91.8 & 94.8 & 84.8 & 77.2 & 87.15 & 38.1 \\
$h=10$ & 89.8 & 92.8 & 88.0 & 79.2 & 87.45 & 19.2 \\
$h=30$ & 68.4 & 72.4 & 79.6 & 69.8 & 72.55 & 7.8 \\
$h=50$ & 53.2 & 55.0 & 72.4 & 46.6 & 56.80 & 5.4 \\
\midrule
AutoHorizon & \underline{95.6} & \underline{97.2} & \textbf{92.4} & \textbf{92.6} & \underline{94.45} & 12.7 \\
\textbf{PAV (Ours)} & \textbf{96.8} & \textbf{98.8} & \underline{91.8} & \underline{91.2} & \textbf{94.65} & 14.0 \\
\midrule[0.9pt]
\end{tabular}
\end{table*}

\paragraph{Learning local validity from trajectory outcomes.}
Having predicted each prefix's consequence, we assess its compatibility with successful continuation through the verification factor in \Eqref{eq:value-decomposition}. We use the predicted representation to summarize the context for this conditional judgment and train a verifier $q_\phi$ from task outcomes. Its output $q_t(h)\in(0,1)$ at $\hat z_{t+h\mid t}$ estimates local validity. To accumulate these conditional estimates into a prefix score, we follow \Eqref{eq:validity-recursion}, starting from $\hat S_t(0)=1$,
\begin{equation}
\begin{aligned}
q_t(h)
&=\frac{\hat S_t(h)}{\hat S_t(h-1)}
=q_\phi(\hat z_{t+h\mid t}),\; 
\hat S_t(h)=\prod_{j=1}^{h}q_t(j),
\end{aligned}
\label{eq:prefix-survival}
\end{equation}
Thus $q_\phi(f_\theta(z_t,\mathbf A_t^{1:h}))$ implements the prediction--verification approximation, and its accumulated product estimates $V_t(h)$ without assuming independent extensions.

Terminal outcomes provide trajectory-level evidence rather than direct labels for local compatibility. We therefore extend score accumulation across policy queries and use these outcomes as surrogate supervision, leaving extension-level validity latent. Let $\mathcal D^+$ and $\mathcal D^-$ collect successful and failed rollouts, with $\mathcal D=\mathcal D^+\cup\mathcal D^-$. For rollout $\tau$, let $\mathcal I_\tau$ index its executed extensions, $n_\tau=|\mathcal I_\tau|$, and $q_i$ denote the corresponding $q_t(h)$, giving $\hat S_\tau=\prod_{i\in\mathcal I_\tau}q_i$. Success supports validity throughout, so we raise every factor, averaging over $n_\tau$ to avoid overweighting long episodes. For failed rollouts, we lower only the product so that the model can retain high scores for compatible extensions without assigning a failure label to each one. This trajectory-level supervision yields
\begin{equation}
  \mathcal L_{\mathrm{traj}}(\tau)
  =\begin{cases}
    -\dfrac{1}{n_\tau}\log\hat S_\tau,
      &\tau\in\mathcal D^+,\\[4pt]
    -\log\bigl(1-\hat S_\tau\bigr),
      &\tau\in\mathcal D^-.
  \end{cases}
  \label{eq:terminal-loss}
\end{equation}
Leaving failure attribution latent avoids labeling every extension as a failure, but the product can still decrease through small validity reductions across the rollout, including at valid extensions. We therefore penalize $\sum_i(1-q_i)$ on failed rollouts. At a fixed product, minimizing this sum favors concentrating the decrease in fewer extensions, allowing high validity elsewhere. With nonnegative weights $\lambda_{\mathrm{traj}}$ and $\lambda_{\mathrm{loc}}$ for outcome supervision and localization, the verification objective becomes
\begin{equation}
  \mathcal L_{\mathrm{ver}}(\theta,\phi)
  = \lambda_{\mathrm{traj}}\,\mathbb E_{\tau\sim\mathcal D}
    [\mathcal L_{\mathrm{traj}}(\tau)]
  + \lambda_{\mathrm{loc}}\,\mathbb E_{\tau\sim\mathcal D^-}
    \biggl[\frac{1}{n_\tau}\sum_{i\in\mathcal I_\tau}(1-q_i)\biggr].
  \label{eq:verification-objective}
\end{equation}
\paragraph{Joint optimization.}
To make predicted representations informative for verification as well as faithful to execution, we jointly minimize $\mathcal L_{\mathrm{pred}}(\theta)+\mathcal L_{\mathrm{ver}}(\theta,\phi)$. Outcome supervision thus reaches $f_\theta$ through the predicted representations used by $q_\phi$. Appendix~\ref{app:training} details the implementation, including label smoothing, minibatch weighting, and auxiliary terminal supervision.

\subsection{Execution-Horizon Selection}
\label{sec:horizon-selection}

We apply length normalization to the accumulated validity score to moderate its dependence on prefix length, obtaining $\bar S_t(h)$ for horizon selection. Appendix~\ref{app:sqrt} provides the formulation. We check candidate prefixes in ascending order and stop extending the horizon at the first score below $\beta\in(0,1)$, giving the execution rule
\begin{equation}
    h_t^\star
    =\max\!\Bigl(\{\min\mathcal H\}\cup
      \Bigl\{h\in\mathcal H\,\Bigm|\,
        \min_{j\in\mathcal H,\,j\le h}\bar S_t(j)
        \ge \beta
      \Bigr\}\Bigr).
  \label{eq:execution-horizon}
\end{equation}
The robot executes the selected prefix and queries the policy from the resulting observation.

\providecommand{\blankresult}{\underline{\hspace{0.55cm}}}
\providecommand{\blankresult}{\underline{\hspace{0.55cm}}}
\providecommand{\notreported}{\textemdash}

\begin{table}[t]
\centering
\setlength{\aboverulesep}{1.5pt}
\setlength{\belowrulesep}{1.5pt}
\begin{minipage}[t]{0.53\textwidth}
\vspace{0pt}
\caption{\small RoboTwin 2.0 with $\pi_{0.5}$, $H=32$ (baseline $h=32$). Success (\%) and mean queries cover DVAC's Selected 16 and All 50 tasks. $\dagger$ marks published results, and dashes mark unreported values.}
\label{tab:robotwin_results}
\end{minipage}\hfill
\begin{minipage}[t]{0.43\textwidth}
\vspace{0pt}
\caption{\small LIBERO ablations with $\pi_{0.5}$, $H=50$. Success is in \%, with overall mean queries. PAV matches the setting and results in Table~\ref{tab:libero_results}.}
\label{tab:ablations}
\end{minipage}
\par\vspace{3pt}
\begin{minipage}[t]{0.53\textwidth}
\vspace{0pt}
\footnotesize
\renewcommand{\arraystretch}{1.0}
\setlength{\tabcolsep}{0.5pt}
\begin{tabular*}{\linewidth}{@{\extracolsep{\fill}}lcccc@{}}
\toprule
& \multicolumn{2}{c}{Selected 16} & \multicolumn{2}{c}{All 50} \\
\cmidrule(lr){2-3}\cmidrule(l){4-5}
Method & Success\,$\uparrow$ & Queries\,$\downarrow$ & Success\,$\uparrow$ & Queries\,$\downarrow$ \\
\midrule
$h=4$ & 17.38 & 181.3 & 27.40 & 136.5 \\
$h=8$ & 46.75 & 75.0 & 55.76 & 54.5 \\
$h=20$ & 47.88 & 28.2 & 55.16 & 20.8 \\
$h=32$ & 35.38 & 19.7 & 43.64 & 14.6 \\
\midrule
DVAC$^\dagger$ & 41.6 & \notreported & \notreported & \notreported \\
\textbf{PAV (Ours)} & \textbf{54.13} & 59.8 & \textbf{58.64} & 58.5 \\
\bottomrule
\end{tabular*}
\end{minipage}\hfill
\begin{minipage}[t]{0.43\textwidth}
\vspace{0pt}
\footnotesize
\renewcommand{\arraystretch}{1.0}
\setlength{\tabcolsep}{1.5pt}
\begin{tabular*}{\linewidth}{@{\extracolsep{\fill}}lccc@{}}
\toprule
Suite / metric & Direct & DINOv2 & PAV \\
\midrule
Spatial & 83.6 & 85.4 & 96.8 \\
Object & 91.8 & 94.2 & 98.8 \\
Goal & 85.0 & 89.8 & 91.8 \\
Long & 77.8 & 78.6 & 91.2 \\
\midrule
Avg. success $\uparrow$ & 84.55 & 87.00 & 94.65 \\
Avg. queries $\downarrow$ & 22.3 & 22.2 & 14.0 \\
\bottomrule
\end{tabular*}
\end{minipage}
\end{table}

\section{Experiments}
\label{sec:experiments}

\subsection{Experimental Setup}

\begin{wrapfigure}{R}{0.50\textwidth}
\vspace{-6pt}
\centering
\setlength{\abovecaptionskip}{4pt}
\includegraphics[width=\linewidth]{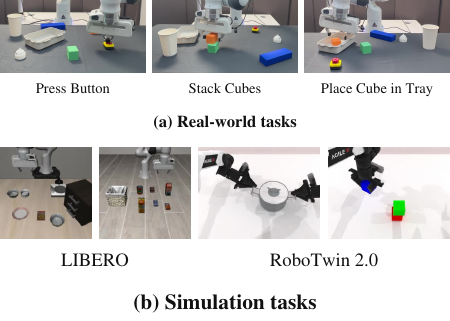}
\caption{\textbf{Experimental settings.} \textbf{(a)} Successful real-world task states for Press Button, Stack Cubes, and Place Cube in Tray, shown from left to right. \textbf{(b)} Representative simulation environments from LIBERO and RoboTwin 2.0, covering single-arm tabletop and dual-arm manipulation.}
\label{fig:experimental_settings}
\vspace{-6pt}
\end{wrapfigure}

\paragraph{Models and Benchmarks.}
We evaluate two representative flow-based VLAs, $\pi_0$ and $\pi_{0.5}$, to test the execution criterion across action-generating policies \citep{black2024pi_0,intelligence2025pi_}. On LIBERO, we use the Spatial, Object, Goal, and Long suites to compare spatial relations, object identity, goal variation, and long-horizon composition in single-arm manipulation \citep{liu2023libero}, excluding LIBERO-90. We evaluate $\pi_0$ with $H=50$ and $\pi_{0.5}$ with $H=10$. To examine a wider range of execution lengths, we also use a $\pi_{0.5}$ checkpoint fine-tuned with $H=50$, although checkpoint differences prevent isolating the effect of chunk length. We then evaluate $\pi_{0.5}$ with its default $H=32$ on all 50 RoboTwin 2.0 tasks to test dual-arm manipulation with varied objects and scenes \citep{chen2025robotwin}. Finally, task-fine-tuned $\pi_{0.5}$ policies with $H=50$ test physical placement, stacking, and contact tasks. Together, these settings assess the criterion across task structures and manipulation setups (Figure~\ref{fig:experimental_settings}).

\paragraph{Training and implementation details.}
To expose the verifier to varied execution lengths, we collect rollouts with randomly sampled horizons and balance successful and failed outcomes during training. In simulation, one verifier is shared across tasks within each policy configuration and benchmark, with evaluation trajectories excluded from training. To isolate method differences, corresponding ablations reuse the same training data, while controlled comparisons share the policy checkpoint and action-generation settings. Appendix~\ref{app:implementation} provides the architecture, data collection and splits, optimization settings, and execution parameters.

\paragraph{Metrics.}
We report success over 50 trials per simulation task and 30 per real-world task alongside mean VLA queries per episode to assess performance relative to inference cost. Each action-chunk generation counts as one query. Selected-horizon distributions characterize execution choices, while verifier computation time measures the added cost of horizon selection.

\subsection{Simulation Experiments}

\subsubsection{LIBERO}

Table~\ref{tab:libero_results} compares our method with fixed horizons, AutoHorizon, and $A^3$ using their public implementations \citep{wang2026vla,chen2026dynamic}. Published DVAC and VLA-Corrector results, marked with $\dagger$, provide additional references for generation stability and prediction agreement \citep{feng2026denoising,pan2026vla}.

Our method achieves the highest average success across all three policy--chunk settings. The clearest gain over fixed execution occurs with $\pi_{0.5}$ at $H=50$, where it improves success by 7.50 percentage points over $h=5$ while reducing queries by 63.3\%. At $H=10$, it raises success from 97.50\% to 98.10\% with similar query counts, showing benefits even with fewer execution choices. Appendices~\ref{app:threshold} and~\ref{app:sqrt} examine comparisons at similar query counts and the effect of length normalization, respectively.

\subsubsection{RoboTwin 2.0}

Table~\ref{tab:robotwin_results} compares our method with fixed horizons $h\in\{4,8,20,32\}$ across all 50 tasks and DVAC's 16-task subset for comparison with its published results \citep{feng2026denoising}. Appendix~\ref{app:robotwin_results} provides per-task results for the Selected 16 tasks.

Our method achieves 54.13\% success on the selected 16 tasks, exceeding the best fixed horizon (47.88\%) and DVAC (41.6\%). This advantage extends to all 50 tasks, where it reaches 58.64\% versus 55.76\% for the best fixed horizon ($h=8$), with similar query counts (58.5 versus 54.5).

\subsubsection{Analysis of Prediction and Policy Alignment}

We first test whether the predictor captures the effects of proposed actions, then compare our method against verification without future prediction (Direct) and prediction in DINOv2 feature space to assess their effects on task success and policy-query cost (Table~\ref{tab:ablations}). We conduct these experiments on the four LIBERO suites using $\pi_{0.5}$ with $H=50$, where ten candidate execution lengths allow us to examine horizon selection over a wider range than the two choices available with $H=10$.

\paragraph{Does the predictor capture action consequences?}
For a predicted representation to inform proposal evaluation, it must capture the effects of the proposed actions. On held-out rollouts, we therefore compare predicted and observed future latents while holding the current input and observed target fixed. Replacing the executed actions with another same-task sequence tests sensitivity to action content. Table~\ref{tab:action_prediction} compares predictions conditioned on executed, same-task swapped, reversed, and zero actions. The prediction conditioned on the executed actions achieves a cosine similarity of 0.9401, compared with 0.5304 for swapped actions.

\begin{table}[t]
\centering
\begin{minipage}[t]{0.46\textwidth}
\vspace{0pt}
\caption{Future-prediction cosine similarity on held-out $\pi_{0.5}$, $H=50$ LIBERO rollouts, with the context and target held fixed.}
\label{tab:action_prediction}
\end{minipage}\hfill
\begin{minipage}[t]{0.50\textwidth}
\vspace{0pt}
\caption{Action-generation and verifier GPU compute times in ms. Parentheses give verifier time relative to generation.}
\label{tab:overhead}
\end{minipage}
\par\vspace{5pt}
\begin{minipage}[t]{0.46\textwidth}
\vspace{0pt}
\footnotesize
\renewcommand{\arraystretch}{1.12}
\setlength{\tabcolsep}{3pt}
\begin{tabular*}{\linewidth}{@{\extracolsep{\fill}}lc@{}}
\toprule
Prediction input & Cosine $\uparrow$ \\
\midrule
Executed actions & 0.9401 \\
Same-task swapped actions & 0.5304 \\
Reversed actions & 0.4816 \\
Zero actions & 0.4103 \\
\bottomrule
\end{tabular*}
\end{minipage}\hfill
\begin{minipage}[t]{0.50\textwidth}
\vspace{0pt}
\footnotesize
\renewcommand{\arraystretch}{1.12}
\setlength{\tabcolsep}{2pt}
\begin{tabular*}{\linewidth}{@{\extracolsep{\fill}}lcc@{}}
\toprule
Setting & Generation & Verifier \\
\midrule
$\pi_0$, LIBERO $H=50$       & 180.34 & 2.24 (1.24\%) \\
$\pi_{0.5}$, LIBERO $H=10$   & 201.19 & 0.90 (0.45\%) \\
$\pi_{0.5}$, LIBERO $H=50$   & 216.95 & 2.49 (1.15\%) \\
$\pi_{0.5}$, RoboTwin $H=32$ & 192.20 & 2.08 (1.08\%) \\
\bottomrule
\end{tabular*}
\end{minipage}
\end{table}

\paragraph{Does explicit prediction improve execution?}
To test whether accurate prediction translates into better execution, we compare against Direct, which scores prefixes from current policy representations and actions without predicting future representations. Rollout data, outcome supervision, score accumulation, and horizon selection are held constant. Explicit prediction raises success from 84.55\% to 94.65\% while reducing mean queries from 22.3 to 14.0 (Table~\ref{tab:ablations}), supporting its role in execution decisions.

\paragraph{Does the prediction space matter?}
To test whether policy alignment contributes beyond future prediction, we replace policy representations with DINOv2 features while retaining the prediction--verification pipeline, action-generating policy, and evaluation conditions. DINOv2 achieves 87.00\% success with 22.2 queries per episode (Table~\ref{tab:ablations}). The higher success and lower query cost in policy space support retaining task-conditioned information for proposal evaluation.

\paragraph{Computational overhead.}
These gains must also be considered alongside the computation added by predictive verification. Across the four policy and chunk-length settings in Table~\ref{tab:overhead}, verifier GPU computation takes 0.90--2.49\,ms, or 0.45--1.24\% of action-generation time. These measurements quantify GPU computation rather than end-to-end control-loop latency.

\subsection{Real-World Experiments}

\subsubsection{Experimental Setup}

\paragraph{Environment.}
We use a 7-DoF Franka Panda with front, wrist, and right-side RGB observations to provide scene-level and near-gripper views. For each task, we fine-tune $\pi_{0.5}$ from its base checkpoint on task-specific teleoperated demonstrations, then collect robot rollouts to train the verifier under the common protocol. Appendix~\ref{app:real_world} provides further hardware and training details.

\paragraph{Tasks.}
The three tasks vary the precision required after transport or approach, allowing us to examine execution under different interaction demands (Figure~\ref{fig:experimental_settings}(a)). \emph{Place Cube in Tray} tests transport and placement of an orange cube, while \emph{Stack Cubes} requires placing it on a green cube and makes terminal alignment more restrictive. \emph{Press Button} tests targeted end-effector contact with a red emergency button. To evaluate beyond a single starting arrangement, object positions are randomized within a $50\,\mathrm{cm}\times50\,\mathrm{cm}$ workspace while remaining visible from the wrist camera.

\subsubsection{Real-World Results}

Our method achieves the highest success on all three real-world tasks, averaging 74.4\% compared with 41.1\% for the strongest fixed-horizon baseline, $h=10$ (Table~\ref{tab:real_results}). It also reduces mean replanning counts from 36.54 to 12.87 per episode. Compared with $h=30$, which has a similar replanning count of 11.97, adaptive execution improves average success from 40.0\% to 74.4\%, showing the benefit of varying execution lengths rather than using a constant horizon.

\begin{table}[t]
\centering
\caption{Real-world success rates (\%), average replanning counts, and average execution horizons. Avg. success is the mean across the three tasks. Bold and underlining indicate the highest and second-highest success rates, including ties.}
\label{tab:real_results}
\small
\setlength{\tabcolsep}{3pt}
\renewcommand{\arraystretch}{1.15}
\begin{tabular*}{\linewidth}{@{\extracolsep{\fill}}lcccccc@{}}
\toprule
Method & \shortstack{Place Cube\\in Tray $\uparrow$} & \shortstack{Stack\\Cubes $\uparrow$} & \shortstack{Press\\Button $\uparrow$} & \shortstack{Avg.\\success $\uparrow$} & \shortstack{Avg.\\replans $\downarrow$} & \shortstack{Avg.\\horizon} \\
\midrule
$h=5$  & 16.7 & 10.0 & 23.3 & 16.7 & 69.58 & 5.00 \\
$h=10$ & \underline{66.7} & 16.7 & 40.0 & \underline{41.1} & 36.54 & 10.00 \\
$h=30$ & 23.3 & \underline{43.3} & \underline{53.3} & 40.0 & 11.97 & 30.00 \\
$h=50$ & 26.7 & 33.3 & 50.0 & 36.7 & 6.91 & 50.00 \\
\textbf{PAV (Ours)} & \textbf{80.0} & \textbf{66.7} & \textbf{76.7} & \textbf{74.4} & 12.87 & 30.22 \\
\bottomrule
\end{tabular*}
\end{table}

\section{Conclusion}
\label{sec:conclusion}

Effective action-chunk execution depends not only on generating coherent actions, but on deciding how long to commit to them. Our approach uses future prediction to inform this commitment, evaluating proposed execution in the same task-conditioned representation space that guides action generation. Across simulated and real-world manipulation, the results support learning execution decisions from task outcomes rather than relying on consistency alone. More broadly, predictive verification offers a way to improve the use of an existing VLA policy by anticipating the consequences of its proposals before acting on them.

\newpage

\subsection*{AI use statement}

In this work, we used generative AI tools to assist with the implementation of methods after the methodology had been independently developed by the authors. We also used generative AI tools to draft parts of the manuscript, improve its readability and presentation, and check grammar. We did not use generative AI tools to develop the research methodology, formulate hypotheses, or interpret experimental results. All AI-assisted text and code were reviewed, revised, and verified by the authors. We take responsibility for the final content of this work, including all text, claims, and artifacts produced with the aid of generative AI.

\section*{Reproducibility Statement}
To support reproducibility, we will submit our implementation as supplementary material. The Method section details the proposed approach and training objectives, while the appendix provides the architecture, experimental configurations, and implementation details needed to reproduce our experiments.

\bibliography{iclr2027_conference}
\bibliographystyle{iclr2027_conference}

\clearpage
\appendix

\section{Implementation and Configuration}
\label{app:implementation}

We detail the $\pi_{0.5}$ LIBERO configuration with a 50-action chunk, no separate text or task embedding, and acceptance threshold $\beta=0.9$. These settings describe the implementation used for the corresponding analysis setting, rather than assuming an identical architecture or configuration for every policy in the main experiments.

\subsection{Architecture of the Predictive Verifier}
\label{app:architecture}

\paragraph{Policy representation.}
To retain the context used to generate a proposal, we extract the final prefix hidden states from the policy's PaliGemma language model, before action denoising. Masked mean pooling, masked max pooling, and the last non-padding token each provide a 2,048-dimensional vector. Concatenating these vectors with the policy-preprocessed 32-dimensional state yields a 6,176-dimensional context. The VLA parameters remain unchanged, and the instruction enters through its language tokens rather than through a separate verifier input. In particular, disabling the task embedding removes the additional learned task-ID lookup, not the policy's instruction conditioning.

Predicting this context directly would require a substantially wider dynamics model. We therefore standardize it using training-set statistics and learn a projector from 6,176 to 128 dimensions, followed by $\ell_2$ normalization. The predicted and target latents in the method are instantiated in this projected policy-feature space. Current and observed future contexts share the projector, while the future target is detached in the prediction losses. Thus, the policy feature extractor is unchanged, but its downstream projection is learned jointly with prediction and verification.

\paragraph{Action-conditioned prediction.}
To evaluate all ten candidate horizons in one recurrent rollout, the predictor divides the proposed $50\times7$ action sequence into ten blocks of five actions. Each block is standardized using training-set action statistics, flattened to 35 dimensions, and embedded into 128 dimensions. Starting from the projected current context and a zero-initialized recurrent hidden state, a GRU cell combines the current latent with the next action embedding. A residual transition head maps its 192-dimensional hidden state to a 128-dimensional update, which is added to the current latent and normalized to obtain the next prediction. This prediction becomes the input latent for the following block. Teacher forcing is disabled during both training and deployment.

\paragraph{Local verification.}
The local verifier reads only the predicted 128-dimensional future latent and applies a sigmoid to its scalar output to obtain $q$. It receives neither a separate action embedding nor the GRU hidden state, so action information reaches this score through the predicted consequence. Table~\ref{tab:app_architecture} specifies the modules. An additional completion head reads the projected observed terminal context during training and is not used to select horizons.

\begin{table}[htbp]
\centering
\caption{Predictive-verifier architecture for $\pi_{0.5}$ on LIBERO with $H=50$. Widths denote linear-layer dimensions. LN denotes LayerNorm. Dropout is zero throughout.}
\label{tab:app_architecture}
\small
\begin{tabular}{p{0.25\linewidth}p{0.53\linewidth}r}
\toprule
Module & Layers & Parameters \\
\midrule
Context projector & LN, $6176\!\to\!256$, GELU, $256\!\to\!128$, $\ell_2$ normalization & 1,626,560 \\
Action encoder & LN, $35\!\to\!128$, GELU, $128\!\to\!128$, GELU & 21,190 \\
Recurrent transition & GRUCell, input 256, hidden 192 & 259,200 \\
Residual update & LN, $192\!\to\!192$, GELU, $192\!\to\!128$ & 62,144 \\
Local verifier & LN, $128\!\to\!192$, GELU, $192\!\to\!1$, sigmoid & 25,217 \\
Completion head & LN, $128\!\to\!128$, GELU, $128\!\to\!1$ & 16,897 \\
\midrule
Total & Including the training-only completion head & 2,011,208 \\
\bottomrule
\end{tabular}
\end{table}

\subsection{Training Data and Optimization}
\label{app:training}

\paragraph{Rollout collection and splitting.}
To expose the verifier to different execution commitments, we collect 150 rollouts for each of the 40 tasks across the four LIBERO suites, sampling uniformly from $\{5,10,\ldots,50\}$ at every policy query. The policy checkpoint is the $H=50$ LIBERO fine-tuned $\pi_{0.5}$ checkpoint at step 30,000. Observations after each completed five-action block provide future targets, while the terminal task outcome labels the rollout. Unexecuted blocks and incomplete final blocks are masked out rather than assigned local labels. Future-context extraction uses the same policy prefix encoder as the current context.

To prevent repeated initializations from crossing data partitions, we split initial-state IDs separately within each task and keep all rollouts from an assigned initialization together. With seed 7, this yields 4,200 training, 840 validation, and 960 held-out test rollouts from the 6,000 collected episodes. Normalization statistics are computed only from the training partition. Context statistics include its current, observed future, and terminal contexts, while action statistics are computed per action dimension. Standard deviations below $10^{-5}$ are replaced by one. Training samples episodes with replacement using inverse class-frequency weights, so success and failure have equal sampling mass.

\paragraph{Prediction and outcome supervision.}
Future alignment compares each predicted latent with the detached projection of the corresponding observed future. To require sensitivity to the proposed actions, action-swap ranking cyclically permutes action chunks among decision records from the same task within a minibatch. Records without another same-task record are excluded from this comparison. Both predicted representations are compared against the original record's future target, with margin $0.1$. The substituted chunk receives no task-failure label.

The trajectory objective treats successful and failed rollouts differently to retain useful local behavior within failures. In this implementation, a successful rollout uses the mean blockwise binary cross-entropy with a smoothed positive target of $0.95$, while a failed rollout uses the negative log-probability of at least one incompatible extension. The minibatch objective averages the success and failure class means when both classes are present. Localization penalizes $1-q$ over all observed blocks from failed rollouts in the minibatch. This implementation uses a block-weighted average, rather than first averaging within each failed rollout and then weighting rollouts equally.

\paragraph{Auxiliary terminal supervision.}
Because future alignment uses a learned projection, an auxiliary completion objective also trains that representation to retain outcome-relevant information. The completion head takes the projected \emph{observed terminal context} and predicts the rollout's binary success label using class-balanced binary cross-entropy. Its loss is weighted by $0.25$ and added to the prediction and verification objectives during joint optimization. It does not predict continuation success from candidate prefixes, and neither its output nor an additional completion factor enters the execution score. The future-contrastive and factorized-completion objectives available in the code are disabled.

\begin{table}[htbp]
\centering
\caption{Training settings for the no-task-embedding $\pi_{0.5}$ LIBERO $H=50$ verifier.}
\label{tab:app_training}
\small
\begin{tabular}{p{0.48\linewidth}p{0.43\linewidth}}
\toprule
Setting & Value \\
\midrule
Collection and training seed & 7 \\
Rollouts per task / total & 150 / 6,000 \\
Train / validation / test rollouts & 4,200 / 840 / 960 \\
Batch size & 8 episodes, including their decision records \\
Optimizer & AdamW, constant learning rate $3\times10^{-4}$ \\
Training epochs / selected epoch & 100 / 7 \\
Checkpoint selection & Lowest validation trajectory loss \\
Weight decay / gradient norm limit & $10^{-4}$ / 5 \\
Alignment / action-swap weights & 1 / 1 \\
Action-swap margin & 0.1 \\
Trajectory / localization weights & 1 / 0.01 \\
Auxiliary completion weight & 0.25 \\
Successful-block training target & 0.95 \\
Dropout / teacher-forcing probability & 0 / 0 \\
Separate text or task embedding & None \\
\bottomrule
\end{tabular}
\end{table}

\subsection{Candidate Horizons and Execution}
\label{app:execution}

To span each chunk with a regular set of execution choices, we use $\{5,10\}$ for $H=10$, $\{4,8,12,16,20,24,28,32\}$ for $H=32$, and $\{5,10,15,20,25,30,35,40,45,50\}$ for $H=50$.

The five-action prediction stride gives candidate horizons $h_k=5k$ for $k\in\{1,\ldots,10\}$. Here $k$ counts predicted extensions, whereas $h_k$ counts robot control steps. To prevent the accumulated score from favoring short prefixes solely through repeated multiplication, deployment applies square-root length normalization to the number of blocks. Writing $q_{t,j}$ for the local score of block $j$, the score checked at $h_k$ is
\begin{equation}
  \widetilde S_t(h_k)
  =\exp\left(\frac{1}{\sqrt{k}}\sum_{j=1}^{k}\log q_{t,j}\right).
  \label{eq:app_log_acceptance}
\end{equation}
The implementation clips probabilities to $[10^{-12},1]$ before taking logarithms. It scans candidates in ascending order and stops extending the execution horizon at the first score below $0.9$. This intermediate-prefix check prevents acceptance beyond an earlier violation, since the normalized score need not be monotone.

The checkpoint's post-hoc bag calibration is not applied to these local scores, and the alternative score obtained by dividing each $q$ by the training target $0.95$ is disabled. Table~\ref{tab:app_candidates} summarizes the resulting deployment configuration.

\begin{table}[htbp]
\centering
\caption{Deployment settings for the $H=50$ LIBERO configuration.}
\label{tab:app_candidates}
\small
\begin{tabular}{p{0.43\linewidth}p{0.48\linewidth}}
\toprule
Setting & Value \\
\midrule
Generated action chunk & $50\times7$ \\
Prediction stride & 5 control steps \\
Candidate horizons & $5,10,15,20,25,30,35,40,45,50$ \\
Acceptance threshold & $\beta=0.9$ \\
Length normalization & Square root of the block count \\
Candidate check & Stop at the first threshold violation \\
Bag calibration / target normalization & Disabled / disabled \\
\bottomrule
\end{tabular}
\end{table}

\section{Detailed RoboTwin Results}
\label{app:robotwin_results}

To show how the aggregate results vary across tasks, Table~\ref{tab:app_robotwin_tasks} expands the Selected 16 comparison in Table~\ref{tab:robotwin_results} to the complete task list reported in DVAC Table~9 \citep{feng2026denoising}. Our fixed-horizon baselines and verifier use $\pi_{0.5}$ with $H=32$, with 50 evaluation episodes per task. The RoboTwin verifier uses $\beta=0.91$ and includes a task embedding, unlike the LIBERO verifier configuration detailed above. The $h=32$ baseline comes from a separate evaluation, so episode-level initial conditions are not guaranteed to be paired with the other methods. DVAC is included as an external reference under its published $H=50$ policy and evaluation protocol.

\begin{table}[htbp]
\centering
\setlength{\belowcaptionskip}{5pt}
\caption{Per-task success rates (\%) on all 16 RoboTwin tasks in the Selected 16 subset. DVAC$^\dagger$ reproduces the $\alpha=2$ results from its Table~9 \citep{feng2026denoising}, including the reported average rounded to one decimal place. Other averages are computed over 800 episodes. Bold indicates the highest reported value in each row, including ties.}
\label{tab:app_robotwin_tasks}
\small
\renewcommand{\arraystretch}{1.15}
\setlength{\tabcolsep}{6pt}
\begin{tabular*}{\textwidth}{@{\extracolsep{\fill}}lcccccc@{}}
\toprule
Task & $h=4$ & $h=8$ & $h=20$ & $h=32$ & DVAC$^\dagger$ & PAV (Ours) \\
\midrule
Blocks Ranking RGB & 0 & 40 & 62 & 38 & 45 & \textbf{64} \\
Blocks Ranking Size & 2 & 34 & 26 & 14 & 37 & \textbf{50} \\
Hanging Mug & 0 & 6 & \textbf{16} & 6 & 1 & 8 \\
Move Stapler Pad & 14 & 30 & 22 & 14 & 11 & \textbf{42} \\
Open Laptop & 44 & 62 & 84 & 60 & \textbf{90} & 70 \\
Open Microwave & 4 & 26 & 46 & 58 & \textbf{67} & 42 \\
Pick Diverse Bottles & 40 & \textbf{62} & 38 & 36 & 55 & 44 \\
Place A2B Left & 34 & 62 & 64 & 64 & 42 & \textbf{74} \\
Place A2B Right & 22 & 74 & 70 & 56 & 30 & \textbf{84} \\
Place Burger Fries & 44 & \textbf{76} & 56 & 58 & 58 & \textbf{76} \\
Place Dual Shoes & 16 & 48 & 48 & 24 & 27 & \textbf{56} \\
Place Empty Cup & 26 & 50 & 54 & 58 & \textbf{68} & 56 \\
Place Object Basket & 26 & 60 & \textbf{68} & 28 & 51 & 58 \\
Scan Object & 4 & 40 & 34 & 12 & 20 & \textbf{42} \\
Stack Blocks Three & 0 & 40 & 22 & 10 & 13 & \textbf{48} \\
Stack Bowls Three & 2 & 38 & \textbf{56} & 30 & 50 & 52 \\
\midrule
Selected 16 & 17.38 & 46.75 & 47.88 & 35.38 & 41.6 & \textbf{54.13} \\
\bottomrule
\end{tabular*}
\end{table}

\section{Additional Ablations and Analysis}
\label{app:ablations}

The detailed analysis is organized around $\pi_{0.5}$ with $H=50$ on LIBERO-Spatial, LIBERO-Object, LIBERO-Goal, and LIBERO-Long. With our candidate sets, $H=10$ allows only $h\in\{5,10\}$, whereas $H=50$ offers ten choices from 5 to 50 steps. We use this broader range to examine how the verifier changes execution length and how each component affects that choice. Only the overhead measurements span multiple policy settings. Comparisons use the same policy checkpoint, rollout dataset, task set, and evaluation protocol. These controlled ablations characterize the components in the $H=50$ LIBERO setting, complementing the broader evaluation of the complete method.

\subsection{Future Prediction and Representation Ablations}
\label{app:representation_ablation}

Table~\ref{tab:ablations} in the main paper reports the complete suite-level comparison of Direct, DINOv2, and PAV. Success and VLA-query counts are shown together to distinguish performance improvements from changes in replanning frequency. Individual loss ablations below further isolate the contribution of the training objectives.

\subsection{Acceptance-Threshold Sweep}
\label{app:threshold}

The threshold changes how much evidence is required before execution and therefore affects both success and query cost. To separate this operating-point choice from representation learning, the sweep holds the trained checkpoint fixed and varies only $\beta$. We evaluate $\beta\in\{0.88,0.89,0.90,0.91,0.92\}$ around the default value of $0.90$.

To compare adaptive and constant-horizon execution at similar query counts, we relate Table~\ref{tab:app_threshold} to the $\pi_{0.5}$, $H=50$ baselines in Table~\ref{tab:libero_results}. At $\beta=0.91$, our method achieves 89.45\% average success with 19.7 VLA queries per episode, compared with 87.45\% and 19.2 queries for $h=10$. This 2.00-percentage-point gain supports an advantage at a similar query budget, complementing the default-setting comparison.

\begin{table}[!htbp]
\centering
\caption{Threshold sweep with a fixed trained verifier on $\pi_{0.5}$ with $H=50$ on LIBERO. Success rates are reported per suite and averaged equally across the four suites. Queries are the total histogram selection count divided by 2,000 evaluation episodes; mean horizon weights each selection equally. The default row uses the PAV results from the main paper.}
\label{tab:app_threshold}
\small
\setlength{\tabcolsep}{3pt}
\renewcommand{\arraystretch}{1.12}
\begin{tabular*}{\linewidth}{@{\extracolsep{\fill}}lccccccc@{}}
\toprule
& \multicolumn{5}{c}{Success (\%) $\uparrow$} & & \\
\cmidrule(lr){2-6}
$\beta$ & Spatial & Object & Goal & Long & Avg. & Queries $\downarrow$ & Mean $h$ \\
\midrule
0.88 & 79.4 & 87.2 & 84.6 & 79.8 & 82.75 & 13.8 & 15.200 \\
0.89 & 81.8 & 90.0 & 85.4 & 85.4 & 85.65 & 15.1 & 13.443 \\
Default (0.90) & 96.8 & 98.8 & 91.8 & 91.2 & 94.65 & 14.0 & 12.783 \\
0.91 & 89.4 & 94.6 & 87.6 & 86.2 & 89.45 & 19.7 & 9.685 \\
0.92 & 90.4 & 94.8 & 86.8 & 81.2 & 88.30 & 25.1 & 7.614 \\
\bottomrule
\end{tabular*}
\end{table}

\begin{figure}[!htbp]
\centering
\includegraphics[width=\linewidth]{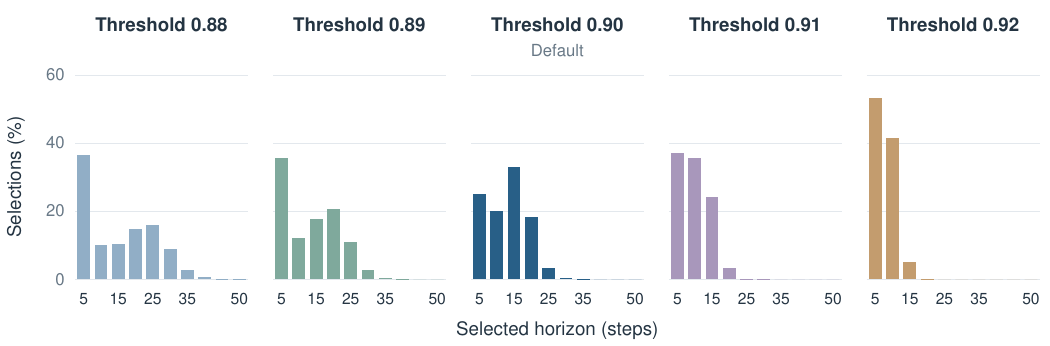}
\caption{Selected-horizon distributions for $\pi_{0.5}$ with $H=50$ on LIBERO. Each panel shows one threshold, with counts pooled across the four suites and normalized within that setting. Panels share the same axes, with one bar per candidate horizon. The default setting is highlighted in dark blue.}
\label{fig:app_threshold_horizons}
\end{figure}

To explain how the threshold changes query demand, Figure~\ref{fig:app_threshold_horizons} compares the selected-horizon distributions across the five sweep settings. We normalize the pooled counts within each setting so that differences in the total number of decisions do not obscure the shift in execution length. At the default $\beta=0.90$, $h=15$ is the most frequent choice, accounting for 33.0\% of selections. The two highest thresholds concentrate execution more strongly on short prefixes, with $h\le10$ accounting for 72.6\% of selections at 0.91 and 94.7\% at 0.92. The default threshold achieves the highest success among the tested settings (Table~\ref{tab:app_threshold}), while higher thresholds shift selections toward shorter prefixes.

\subsection{Removing Individual Loss Terms}
\label{app:loss_ablation}

To determine which training constraints contribute to useful execution decisions, each variant removes one term while retaining the architecture, data, and remaining coefficients. Removing future alignment sets $\lambda_{\mathrm{dyn}}=0$ to isolate the contribution of observed future targets. Removing swap ranking sets $\lambda_{\mathrm{swap}}=0$ and tests whether alignment encourages sufficient action sensitivity. Removing localization sets $\lambda_{\mathrm{loc}}=0$ to evaluate how the localization prior contributes to execution performance.

The trajectory objective is retained to provide a common source of outcome supervision. Each trained variant is evaluated with the default threshold, so the comparison measures the contribution of each term under the same execution rule.

\begin{table}[htbp]
\centering
\caption{Loss-term ablations on $\pi_{0.5}$ with $H=50$ on LIBERO. Suite columns and the average report success (\%). Queries are the mean number of VLA generations per episode across the four suites.}
\label{tab:app_loss_ablation}
\resizebox{\linewidth}{!}{
\begin{tabular}{lrrrrrr}
\toprule
Training variant & Spatial & Object & Goal & Long & Avg. & Queries \\
\midrule
Full objective & 96.8 & 98.8 & 91.8 & 91.2 & 94.65 & 14.0 \\
Without future alignment & 83.2 & 91.0 & 86.8 & 80.2 & 85.30 & 23.99 \\
Without swap ranking & 84.6 & 93.4 & 88.2 & 76.6 & 85.70 & 24.26 \\
Without localization & 81.8 & 92.2 & 86.8 & 73.8 & 83.65 & 25.60 \\
\bottomrule
\end{tabular}
}
\end{table}

The full objective improves success by 8.95--11.00 percentage points over the three ablated variants while reducing mean queries from 23.99--25.60 to 14.0 (Table~\ref{tab:app_loss_ablation}). These results support retaining future alignment, action-swap ranking, and localization in the evaluated configuration.

\subsection{Sensitivity to Action Swaps}
\label{app:action_swap}

Table~\ref{tab:action_prediction} evaluates the trained predictor with executed and same-task swapped actions, alongside reversed and zero-action inputs. We further examine how the effect of replacing the actions varies across prediction horizons. This input-sensitivity diagnostic complements the separately trained loss ablations in Table~\ref{tab:app_loss_ablation}.

A predictor may align with future observations by exploiting scene regularities while responding weakly to the proposed actions. To test this possibility, held-out transitions are evaluated with both the executed prefix and a different prefix of the same length. The current representation and observed future target are held fixed. Each donor comes from a different held-out episode of the same task with the same number of observed blocks. We retain only recorded future targets with an eligible donor, yielding 36,843 observed-block pairs across 960 test episodes. Each pair receives equal weight.

Using the cosine distance from \Eqref{eq:prediction-objective}, define the prediction-distance gap
\begin{equation}
  \Delta_t(h)=d_{\cos}(\tilde z_{t+h\mid t},z_{t+h})
              -d_{\cos}(\hat z_{t+h\mid t},z_{t+h}).
  \label{eq:app_swap_gap}
\end{equation}
A positive gap means that the executed prefix predicts its observed consequence more accurately than the alternative. We report the gap by horizon together with the fractions satisfying $\Delta_t(h)>0$ and $\Delta_t(h)\ge m$. Both inputs are evaluated against the same recorded future, so the comparison measures action sensitivity relative to that target. Across horizons, the executed prefix gives a closer prediction in 89.3--95.2\% of pairs, supporting action sensitivity over the tested prediction range.

\begin{table}[htbp]
\centering
\caption{Horizon-wise action-swap sensitivity of the full predictor on held-out transitions. $n$ is the number of observed-block pairs, and the ranking margin is $m=0.1$.}
\label{tab:app_swap}
\small
\begin{tabular}{ccccc}
\toprule
Horizon & $n$ & Mean $\Delta_t(h)$ & $\Delta_t(h)>0$ (\%) & $\Delta_t(h)\ge m$ (\%) \\
\midrule
5 & 7,276 & 0.1628 & 89.3 & 53.5 \\
10 & 6,367 & 0.3292 & 92.5 & 70.7 \\
15 & 5,537 & 0.4257 & 93.6 & 76.2 \\
20 & 4,683 & 0.4898 & 94.6 & 79.6 \\
25 & 3,883 & 0.5287 & 94.7 & 81.0 \\
30 & 3,160 & 0.5537 & 95.2 & 83.0 \\
35 & 2,470 & 0.5619 & 94.7 & 82.7 \\
40 & 1,781 & 0.5697 & 94.6 & 81.9 \\
45 & 1,140 & 0.5636 & 94.0 & 81.8 \\
50 & 546 & 0.5792 & 95.2 & 84.6 \\
\bottomrule
\end{tabular}
\end{table}

\subsection{Effect of Square-Root Length Normalization}
\label{app:sqrt}
\label{app:horizon_distribution}

Following Appendix~\ref{app:execution}, let $k$ count five-action blocks and $h_k=5k$ denote the execution length. The accumulated score is $\hat S_t(h_k)=\prod_{j=1}^{k}q_{t,j}$, which becomes $q^k$ when local validity is constant. To permit longer prefixes with consistently high local scores while retaining a length penalty, we moderate this decay to $q^{\sqrt k}$ using
\begin{equation}
  \bar S_t(h_k)=\hat S_t(h_k)^{1/\sqrt k}.
  \label{eq:sqrt-normalization}
\end{equation}
This normalized score is used for horizon selection rather than as a calibrated probability. Because normalization can raise later scores after an earlier decrease, the execution rule checks candidate prefixes in ascending order and stops at the first threshold violation. A later increase therefore cannot override an earlier violation.

This ablation tests whether square-root normalization enables longer execution while preserving or improving task success on $\pi_{0.5}$ with $H=50$ across the four LIBERO suites. Because it acts on the accumulated score at inference rather than as a training regularizer, we hold the trained predictor and verifier fixed and change only the normalization rule. We compare raw accumulation, square-root normalization, and geometric-mean normalization, written as
\begin{equation}
  S_t^{(\eta)}(h_k)=\hat S_t(h_k)^{1/k^\eta},
  \qquad \eta\in\{0,\tfrac12,1\}.
  \label{eq:app_length_family}
\end{equation}
Here $\eta=0$ gives the unnormalized product, $\eta=\tfrac12$ gives our rule, and $\eta=1$ gives the geometric mean. For constant local validity $q$, these scores reduce to $q^{k^{1-\eta}}$. Raw accumulation therefore decays exponentially in $k$, while the geometric mean removes length dependence in this constant-score case. Square-root normalization retains a length penalty but moderates its growth.

All three variants retain the intermediate-prefix acceptance check and the same candidate horizons $h\in\{5,10,\ldots,50\}$. Table~\ref{tab:app_sqrt} compares suite-level success, overall query count, and mean horizon at the common default threshold. This comparison evaluates all three rules at the shared operating point.

\begin{figure}[!ht]
\centering
\setlength{\abovecaptionskip}{4pt}
\setlength{\belowcaptionskip}{0pt}
\includegraphics[width=\linewidth]{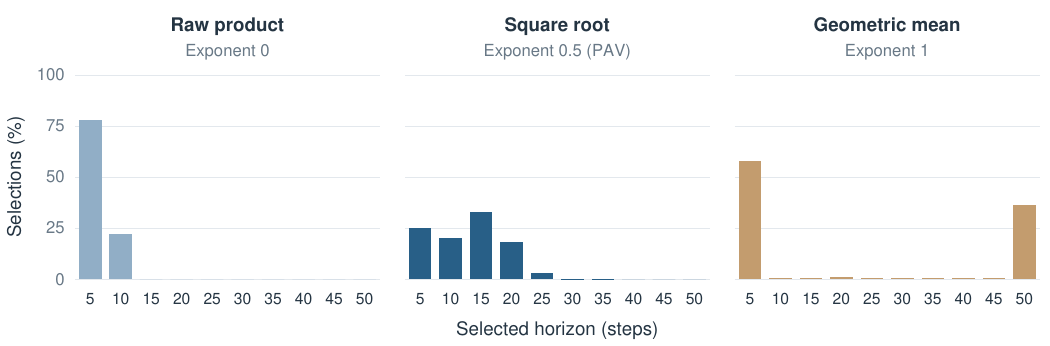}
\caption{Horizon selection under three normalization rules at $\beta=0.90$ on $\pi_{0.5}$, LIBERO $H=50$. Each panel uses identical axes and independently normalized counts pooled across suites. Square-root normalization favors intermediate horizons. Success and query costs appear in Table~\ref{tab:app_sqrt}.}
\label{fig:app_horizon_distribution}
\end{figure}

To explain how normalization changes execution beyond its mean length, Figure~\ref{fig:app_horizon_distribution} shows one selected-horizon histogram per rule, pooling decisions across the four LIBERO suites and normalizing within each variant. Raw accumulation selects only $h=5$ or $h=10$, whereas square-root normalization shifts the mode to $h=15$ and spreads selections across intermediate horizons. Geometric-mean normalization instead concentrates selections at both $h=5$ and $h=50$. At the shared threshold, square-root normalization achieves the highest task success and favors intermediate execution lengths, supporting its use as the default normalization in this setting (Table~\ref{tab:app_sqrt}).

\begin{table}[!ht]
\centering
\caption{Inference-time normalization ablation with the trained predictor and verifier held fixed. Success rates are reported for each LIBERO suite and averaged equally across the four suites. Queries are the total histogram selection count divided by 2,000 evaluation episodes; mean horizon weights each selection equally. Entries correspond to the common default threshold.}
\label{tab:app_sqrt}
\small
\setlength{\tabcolsep}{3pt}
\renewcommand{\arraystretch}{1.12}
\begin{tabular*}{\linewidth}{@{\extracolsep{\fill}}lccccccc@{}}
\toprule
& \multicolumn{5}{c}{Success (\%) $\uparrow$} & & \\
\cmidrule(lr){2-6}
Normalization & Spatial & Object & Goal & Long & Avg. & Queries $\downarrow$ & Mean $h$ \\
\midrule
Raw product ($\eta=0$) & 92.0 & 90.8 & 91.6 & 80.0 & 88.6 & 30.9 & 6.104 \\
Square root ($\eta=\tfrac12$) & 96.8 & 98.8 & 91.8 & 91.2 & 94.65 & 14.0 & 12.783 \\
Geometric mean ($\eta=1$) & 53.6 & 53.0 & 71.8 & 50.4 & 57.2 & 11.3 & 22.645 \\
\bottomrule
\end{tabular*}
\end{table}

\section{Qualitative Demonstrations}
\label{app:qualitative}

\begin{figure}[htbp]
\centering
\includegraphics[width=\linewidth]{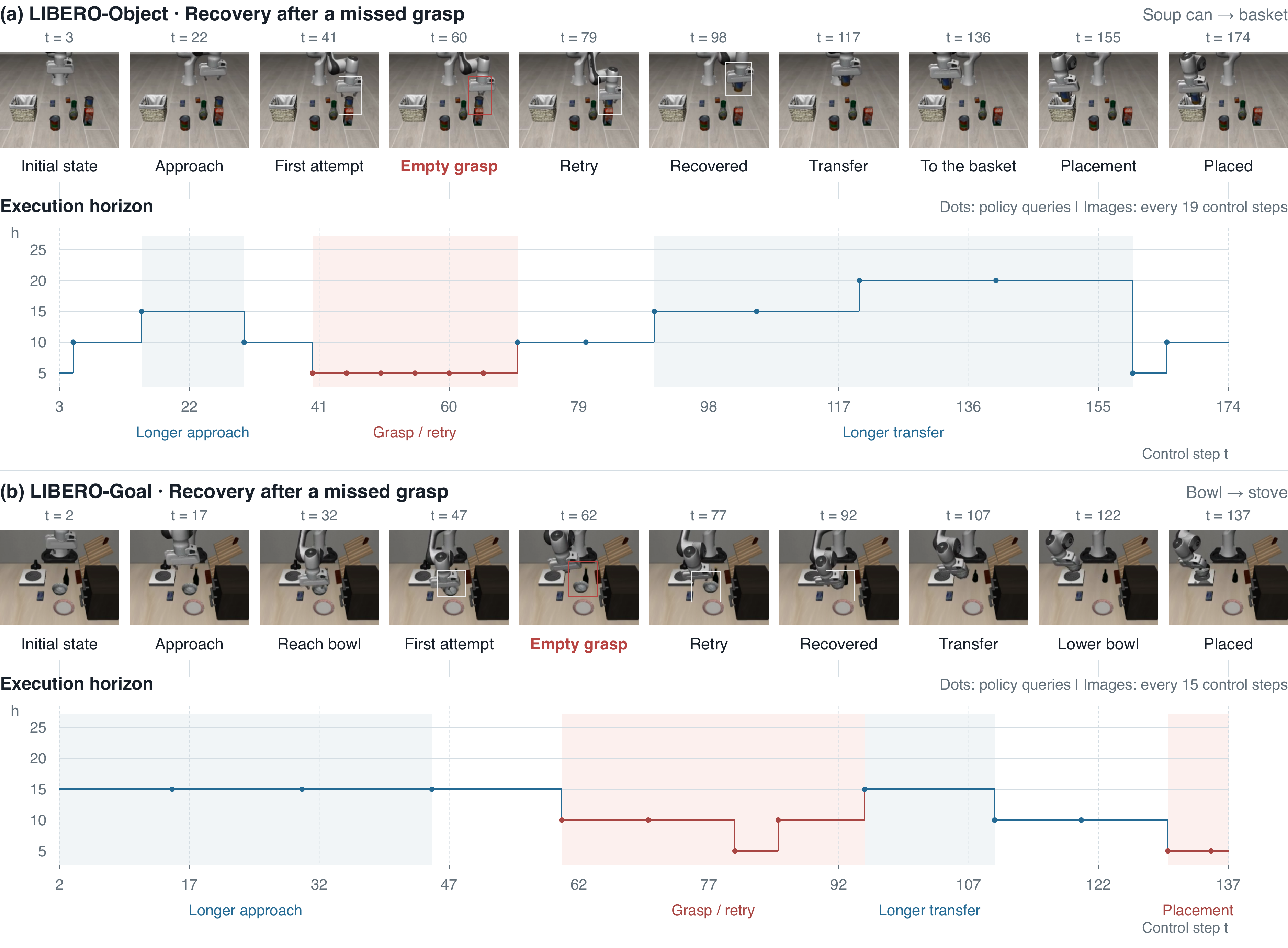}
\caption{Adaptive execution during grasp recovery in two LIBERO rollouts. (a) In LIBERO-Object, the robot retries an empty grasp before transferring the soup can to the basket. (b) In LIBERO-Goal, it reattempts the bowl grasp and places the bowl on the stove. Observed frames are sampled every 19 and 15 control steps, respectively. The traces show selected execution horizons, with dots marking policy queries. Short horizons accompany grasp retries, while longer horizons occur during approach and transfer. Shaded intervals identify the annotated interaction phases, not validity labels or training supervision.}
\label{fig:app_sim_demos}
\end{figure}

To relate horizon selection to physical interactions, Figure~\ref{fig:app_sim_demos} pairs observed rollout frames with execution-horizon traces from LIBERO-Object and LIBERO-Goal. Both examples contain a missed grasp followed by a successful retry and placement. Shorter horizons around the retry provide more frequent opportunities to incorporate feedback, while longer horizons during approach and transfer sustain motion with fewer policy queries. The traces illustrate how execution length changes within a rollout rather than merely differing across tasks.

These examples illustrate shorter commitments during grasp recovery and longer commitments during transfer, complementing the aggregate success and query results. The images show observed rollout frames, and the traces report selected execution horizons.

\clearpage
\section{Real-World Scene and Demonstrations}
\label{app:real_world}

\subsection{Hardware, Views, and Task Layout}

The real-world setup uses a 7-DoF Franka Panda with front, wrist, and right-side RGB observations. Figure~\ref{fig:experimental_settings}(a) shows the gripper and tabletop task layouts at successful task states. Object placements vary within the $50\,\mathrm{cm}\times50\,\mathrm{cm}$ region described in the main text, while remaining visible from the wrist camera at initialization. Policy and verifier inference run on an NVIDIA RTX 4090. Each RGB view is resized with aspect-ratio-preserving padding to $224\times224$ pixels. Actions are executed at a nominal frequency of $15\,\mathrm{Hz}$. The policy interface outputs seven absolute joint-position targets in radians and one gripper command. During training, joint targets are represented as offsets from the current joint configuration and converted back to absolute targets for execution.

\paragraph{Task definitions.}
\emph{Place Cube in Tray} places an orange cube in a tray, \emph{Stack Cubes} places an orange cube on top of a green cube, and \emph{Press Button} activates a red emergency button. A human evaluator judges task success, and each trial has a maximum budget of 500 action steps. Trials terminated early without achieving task success are counted as failures and retained in the success-rate denominator.

\paragraph{Training and evaluation records.}
We fully fine-tune a separate $\pi_{0.5}$ base policy for each task using task-specific teleoperated demonstrations. The deployed policies use the checkpoints obtained after 10,000 optimizer updates, with $H=50$. Each task-specific verifier is trained on 50 rollouts for 60 epochs. Deployment uses candidate horizons $\{5,10,\ldots,50\}$, a common threshold $\beta=0.90$ across all three tasks, and square-root block-count normalization. Each decision evaluates a single generated action chunk and executes its selected prefix without regenerating the proposal. Evaluation uses 30 trials per task, separate from the training data. To match initial conditions across methods, object placements are randomized when constructing the evaluation sequence and then held fixed by episode index, so each method encounters the same initial placement for the corresponding episode.

\subsection{Real-World Execution Examples}

\begin{figure}[p]
\centering
\includegraphics[width=\linewidth]{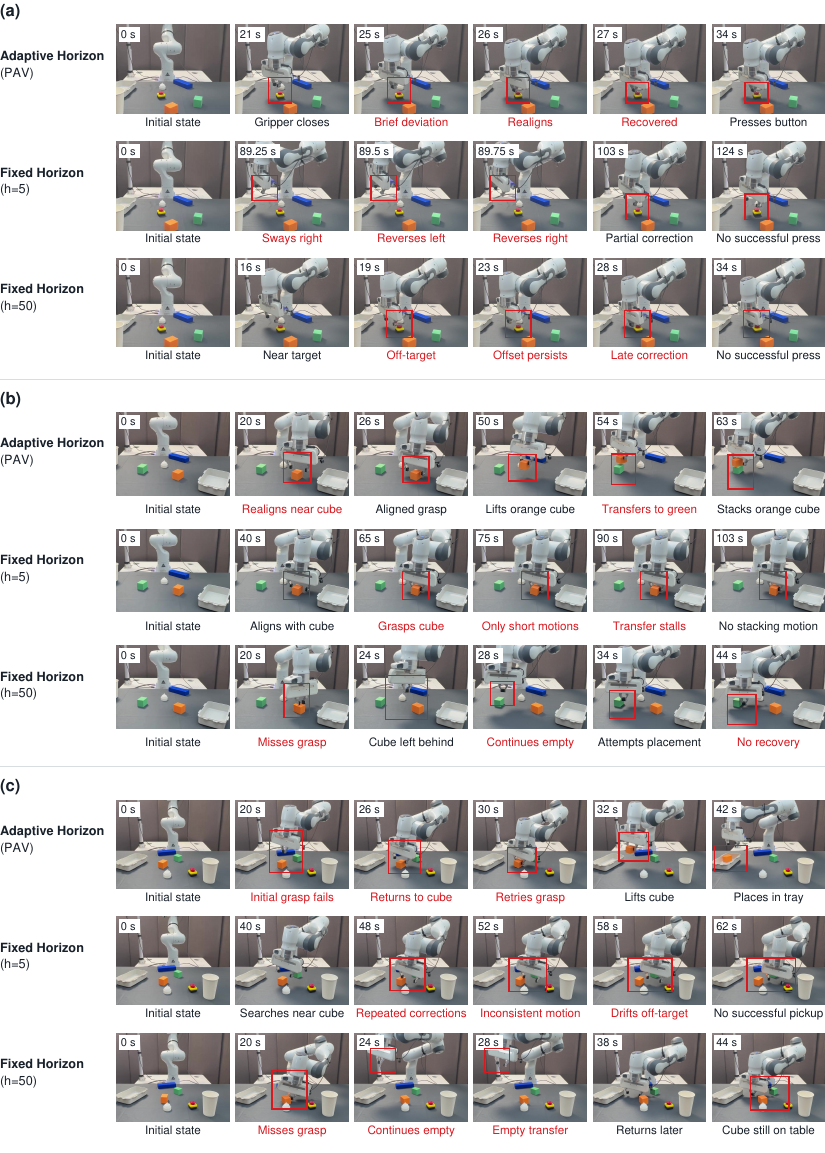}
\caption{\small Real-world qualitative comparisons. \textbf{(a) Press Button:} PAV recovers and presses. $h=5$ oscillates, while $h=50$ retains an off-target alignment. \textbf{(b) Stack Cubes:} PAV realigns and stacks. $h=5$ grasps but stalls before transfer, while $h=50$ continues after a missed grasp. \textbf{(c) Place Cube in Tray:} PAV returns and retries successfully. $h=5$ drifts away, while $h=50$ continues empty before returning later without success. Red annotations highlight key behaviors. All rows start at $t=0$. Video timestamps are nonuniform and not synchronized across rows.}
\label{fig:app_real_demos}
\end{figure}

Figure~\ref{fig:app_real_demos} compares adaptive execution with fixed horizons across three real-world tasks. Adaptive execution combines corrective feedback near interactions with sustained progress afterward. Short execution can interrupt coherent progress, while long execution can continue despite an unsuccessful interaction. Each row shows observations from one rollout; the examples illustrate these behaviors rather than a measured trace of individual policy-query events.

\paragraph{Metric aggregation.}
Replanning counts include the initial action-chunk generation and therefore count total VLA queries per episode. Mean execution horizon is computed over all selected prefixes pooled across evaluation episodes, giving each selection equal weight rather than averaging episode-level means.

\end{document}